\documentclass{article}
\usepackage{spconf,amsmath,graphicx}
\usepackage{array,multirow,booktabs}
\usepackage[dvipsnames]{xcolor}
\usepackage{microtype}
\usepackage{subfigure}
\usepackage{booktabs} 
\usepackage{todonotes}
\usepackage{hyperref}
\usepackage{enumitem}
\DeclareSymbolFont{extraup}{U}{zavm}{m}{n}
\DeclareMathSymbol{\varheart}{\mathalpha}{extraup}{86}
\DeclareMathSymbol{\vardiamond}{\mathalpha}{extraup}{87}

\makeatletter
\newcommand{\thickhline}{%
    \noalign {\ifnum 0=`}\fi \hrule height 1pt
    \futurelet \reserved@a \@xhline
}
\newcolumntype{"}{@{\hskip\tabcolsep\vrule width 1pt\hskip\tabcolsep}}
\makeatother

\title{Looking Enhances Listening: Recovering Missing Speech Using Images }

  \name{Tejas Srinivasan$^{\spadesuit}$, Ramon Sanabria$^{\varheart}$, Florian Metze$^{\spadesuit}$}
\address{
    $^{\spadesuit}$Language Technologies Institute, Carnegie Mellon University, U.S.A\\
	$^{\varheart}$ CSTR and ILCC, University of Edinburgh, UK\\
	\texttt{$\{$tsriniva|fmetze$\}$@cs.cmu.edu, r.sanabria@ed.ac.uk}
	}

\begin{document}
\ninept
\maketitle
\begin{abstract}

Speech is understood better by using visual context; for this reason, there have been many attempts to use images to adapt automatic speech recognition (ASR) systems.
Current work, however, has shown that visually adapted ASR models only use images as a regularization signal, while completely ignoring their semantic content.
In this paper, we present a set of experiments where we show the utility of the visual modality under noisy conditions.
Our results show that multimodal ASR models can recover words which are masked in the input acoustic signal, by grounding its transcriptions using the visual representations. We observe that integrating visual context can result in up to 35\% relative improvement in masked word recovery.
These results demonstrate that end-to-end multimodal ASR systems can become more robust to noise by leveraging the visual context.

\end{abstract}
\section{Introduction}
\label{sec:intro}

Humans process and understand language better when integrating information from multiple modalities. More concretely, in conversations, we use the context that is surrounding us to properly interpret what has been said. Consequently, visual modality integration has recently become a trend in the speech and natural language processing communities. Previous works show improvements in the domains of visual question-answering~\cite{antol2015vqa}, multimodal machine translation~\cite{specia2016shared}, visual dialog~\cite{visdial}, and automatic speech recognition (ASR)~\cite{palaskar2018end}.
Although there are several visual adaption approaches for ASR~\cite{palaskar2018end,Caglayan2018multimodal,moriya2018lstm,gupta2017visual,miao2016open,sun2016look},
it is still unclear how the models leverage the visual modality.





Previous works have demonstrated that the visual modality can be used to individually adapt the language~\cite{gupta2017visual,sun2016look, moriya2018lstm} and acoustic~\cite{miao2016open} model components, as well as end-to-end (e2e) models~\cite{palaskar2018end, srinivasan}. In acoustic model adaptation~\cite{miao2016open}, images can provide acoustic context (\textit{e.g.} outdoors and indoors acoustics can be inferred from images). In a similar vein, breakthroughs in image captioning inspired language models contextualization using visual information \cite{gupta2017visual,sun2016look}. By inferring the domain of a conversation from images, we can bias the language model towards  a desired space and improve the transcription.
However, a lack of analysis and understanding of these models inhibits their widespread application.


In previous work, we analyze the contribution of different visual representations in an end-to-end multimodal ASR system~\cite{Caglayan2018multimodal}. We observe that an ASR model trained with multimodal information is able to maintain its performance even when the visual modality is discarded during test time. This observation indicates that the model is using the visual modality as a regularization technique, and not using the semantics of the image. Similar findings in multimodal machine translation \cite{desmond} led to an investigation of the utility of visual context ~\cite{caglayan2019probing}. They concluded that visual modality is helpful when the primary modality is degraded.


\begin{figure}[t]
    \centering
    \includegraphics[width=0.95\columnwidth]{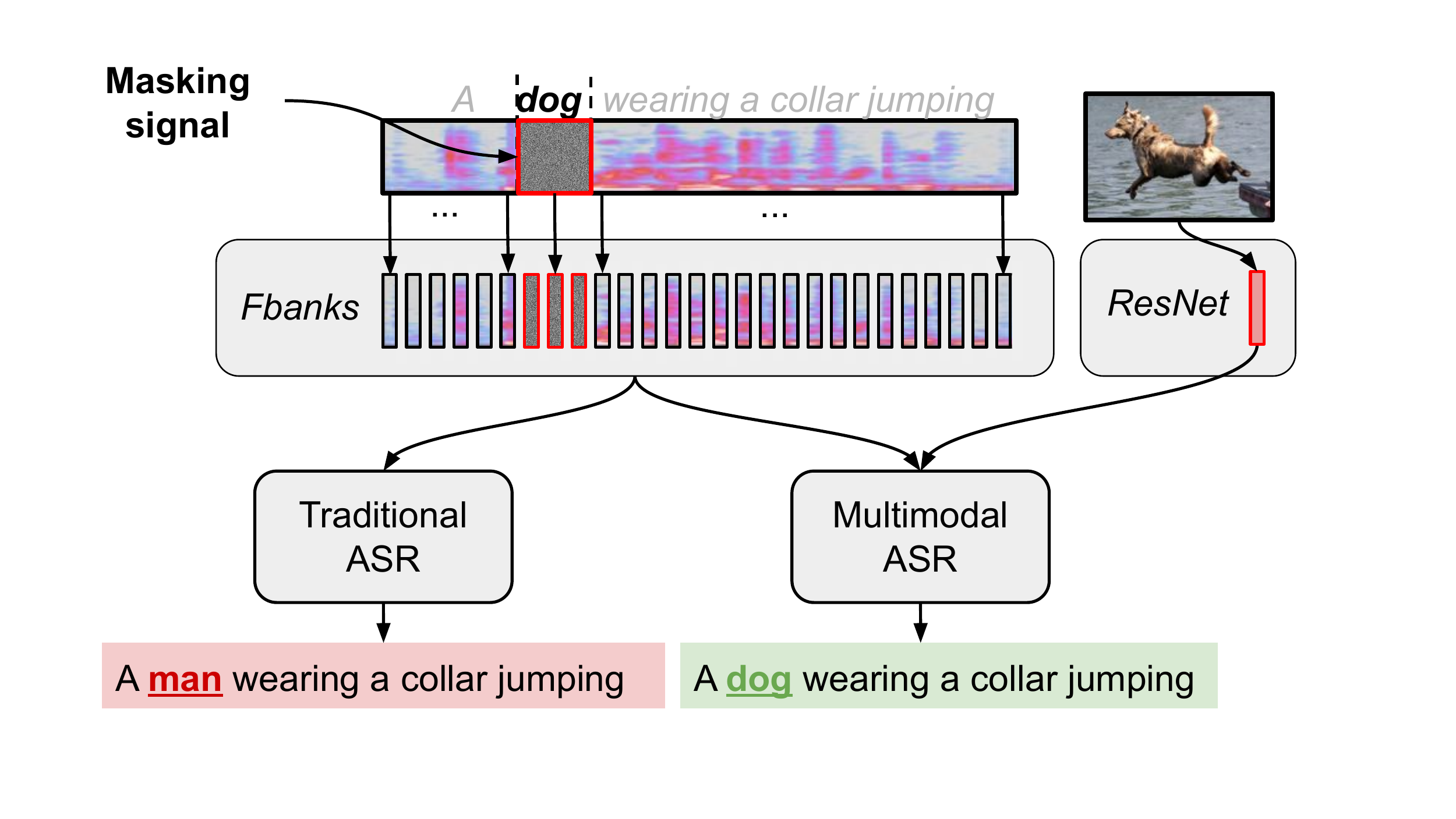}
 \caption{Our set of experiments demonstrate that multimodal speech recognition systems can recover masked words by using images as an auxiliary signal.}
    \label{fig:my_label}
\end{figure}

In this paper, inspired by~\cite{caglayan2019probing}, we hypothesize that Multimodal ASR systems can take advantage of the visual modality under severely noisy conditions and be more robust than unimodal models. Such robustness to noise is particularly desirable in ASR - since speech is inherently a noisy medium.

The main contributions of this paper are as follows:
\begin{itemize}
    \item A set of noise-masking experiments, where we inject different kinds of noise into the acoustic signal in a structured manner (Section \ref{sec:method}).
    \item A comparison of the ability of unimodal and multimodal ASR models to recover the masked audio signal (Section \ref{subsec:results}).
    \item Quantitative (Section \ref{subsec:cong}) and qualitative analysis (Section \ref{subsec:examples}) to investigate how exactly the visual modality is utilized.
\end{itemize}

Our observations on the Flickr 8k Audio Caption Corpus indicate that multimodal ASR models perform better at recovering masked words than unimodal models, thereby validating our hypothesis that multimodal models are more robust to noise. 


Furthermore, our analysis suggests that recovery performance drops when we provide incorrect visual features during training and/or testing - this result indicates that the model is sensitive to the the semantic content of the visual context.








\section{Methodology}
\label{sec:method}


In this section, we introduce our multimodal architecture, similar to~\cite{Caglayan2018multimodal, srinivasan}, that we use to incorporate the image modality in an end-to-end ASR system. We also describe the different masking experiments that we design to test our hypothesis. Inspired by~\cite{caglayan2019probing}, we simulate different scenarios where the audio signal is corrupted by various masking techniques. This audio corruption is done in a controlled manner, by deterministically fixing a set of words and masking all occurrences in the dataset. 


\subsection{Models}

To be consistent with previous work, we test our hypothesis with the best performing models and features according to~\cite{Caglayan2018multimodal, srinivasan}.

\subsubsection{Unimodal ASR}
Our baseline unimodal ASR model is a sequence-to-sequence model with attention~\cite{DBLP:conf/icassp/BahdanauCSBB16, DBLP:conf/icassp/ChanJLV16}. The encoder consists of 6 bidirectional LSTM layers with \textit{tanh} activation, with sub-sampling at the 3rd and 4th layers~\cite{DBLP:conf/icassp/BahdanauCSBB16}. The decoder is a two-layer GRU which computes attention over the encoder states.

\subsubsection{Multimodal ASR}
For our Multimodal ASR model, we use \textbf{early decoder fusion}. In this fusion technique, during decoding, a visual feature vector $f$ is projected down to the hidden state dimension, and further concatenated to the input embedding vector $y_t$ at each timestep before passing it through the GRU decoder.
\begin{align*}
    f' = tanh(W_f f + b_f) ; \quad     y_{t} = [y_t; f']
\end{align*}
\subsection{Audio Corruption}

To create scenarios where information is missing from the primary modality, we perform masking on the audio data. Because we need to have control over the words being masked and recovered, we generate word-level masking. To do so, we first generate forced aligned timings to localize specific words in the transcriptions. After that, we experiment with two masking techniques:
\begin{enumerate}[leftmargin=*]
    \item \textbf{Silence Masking}: We substitute the masked word with a specific value, silence, similar to the special token in~\cite{caglayan2019probing}. In this non-realistic scenario, the model is trained to generate the missing word when a known signal is present in the audio.
    \item \textbf{White Noise Masking}: We substitute the masked word with white noise. This more realistic scenario is an approximation to a noisy-ASR problem where speech is corrupted by some stochastic signal. In this case, in contrast to a traditional noisy-ASR situation where the speech is overlapped with noise, the masked word is completely overwritten.
\end{enumerate}
 It is important to note that the masking is performed on all data splits. This set up differs from~\cite{srinivasan}, where the masking was performed only in the test set and models were trained on unmasked data.

\subsubsection{Masked Words}

To have a more clear insight into how the model uses the image modality, we experiment by masking two different sets of words in the dataset: \textbf{nouns} and \textbf{places}. We analyze the efficiency of the model on recovering these different categories and how the visual representation influences the result. We hypothesize that visual embeddings trained on place recognition should be more effective for place word recovery than objects, and vice versa.

\subsection{Congruency Experiments}
\label{subsec:congruency-exps}

Similar to~\cite{desmond}, we perform a set of experiments where we misalign images and utterances. The outcome of this set up quantifies the sensitivity of our model towards unaligned images. These results, therefore, can provide insights on our previous claim that multimodal models use the visual modality as a regularization signal~\cite{Caglayan2018multimodal}.

\begin{itemize}[leftmargin=*]
    \item \textbf{Incongruent Decoding}: We train multimodal models on the correct (congruent) image-utterance pairings, and deliberately misalign the images and utterances during testing. This is done to check that our multimodal models rely on the images - when presenting an unrelated image, we expect to see poor performance.
    \item \textbf{Incongruent Training}: We both train and test the models on incongruent image-utterance alignments. In this experiment, we wish to see if providing an incorrect visual signal can still result in improvements in the model, similar to the regularization effect observed in \cite{Caglayan2018multimodal}.
\end{itemize}

\section{Experimental Setup}
\label{sec:exp}
\begin{table*}[t]
\label{tab:results}
\centering
\begin{tabular}{|c|c||c|c||c|c|c|c|}
\hline
Masking Type                & Masked Words            & Model      & Visual Features & Recovery Rate & Rel RR $\triangle$ & WER    & Rel WER $\triangle$ \\ \hline \hline
\multirow{3}{*}{None}          & \multirow{3}{*}{-}      & Unimodal   & -               & -             & -                  & 16.4\% & -                   \\ \cline{3-8} 
                            &                         & Multimodal & Object Ftrs     & -             & -                  & 14.8\% & 10.0\%              \\ \cline{3-8} 
                            &                         & Multimodal & Place Ftrs      & -             & -                  & 15.6\% & 4.7\%               \\ \hline \hline
\multirow{3}{*}{Silence}    & \multirow{3}{*}{Nouns}  & Unimodal   & -               & 46.0\%        & -                  & 32.9\% & -                   \\ \cline{3-8} 
                            &                         & Multimodal & Object Ftrs     & 57.8\%        & 25.5\%             & 29.9\% & 4.0\%               \\ \cline{3-8} 
                            &                         & Multimodal & Place Ftrs      & 52.6\%        & 14.2\%             & 31.5\% & 3.2\%               \\ \hline \hline
\multirow{3}{*}{Silence}    & \multirow{3}{*}{Places} & Unimodal   & -               & 42.7\%        & -                  & 22.6\% & -                   \\ \cline{3-8} 
                            &                         & Multimodal & Object Ftrs     & 57.0\%        & 33.5\%             & 19.3\% & 14.6\%              \\ \cline{3-8} 
                            &                         & Multimodal & Place Ftrs      & 53.7\%        & 25.7\%             & 22.0\% & 2.7\%               \\ \hline \hline
\multirow{3}{*}{Whitenoise} & \multirow{3}{*}{Nouns}  & Unimodal   & -               & 43.3\%        & -                  & 43.5\% & -                   \\ \cline{3-8} 
                            &                         & Multimodal & Object Ftrs     & 57.3\%        & 32.3\%             & 36.5\% & 16.2\%              \\ \cline{3-8} 
                            &                         & Multimodal & Place Ftrs      & 50.4\%        & 16.3\%             & 43.1\% & 0.9\%               \\ \hline \hline
\multirow{3}{*}{Whitenoise} & \multirow{3}{*}{Places} & Unimodal   & -               & 42.2\%           & -                  & 25.2\%    & -                   \\ \cline{3-8} 
                            &                         & Multimodal & Object Ftrs     & 57.8\%           & 37.1\%                & 19.3\%    & 23.6\%                 \\ \cline{3-8} 
                            &                         & Multimodal & Place Ftrs      & 53.7\%           & 27.4\%                & 23.5\%    & 7.0\%                 \\ \hline 
\end{tabular}
\caption{Word Error Rate (WER) and Recovery Rate (RR) scores for our unimodal and multimodal models in the different masking conditions. We also show the improvements ($\triangle$s) of our multimodal models, relative to the unimodal model under same masking conditions.}
\end{table*}

\subsection{Dataset}
We perform experiments on the Flickr 8k Audio Caption Corpus~\cite{harwath2015deep}, which contains 40,000 spoken captions (total 65 hours of speech) corresponding to 8,000 natural scene images from the Flickr8k dataset~\cite{DBLP:conf/ijcai/HodoshYH15}\footnote{\url{https://groups.csail.mit.edu/sls/downloads/flickraudio/}}. We use the pre-defined training, development and test splits of 30k, 5k and 5k utterances respectively. The audio and text captions in this data are more structured and clean, compared to the `in-the-wild' nature of How2 \cite{sanabria18how2}.

\subsection{Implementation Details}

\subsubsection{Word Masking}

\begin{itemize}[leftmargin=*]
    \item \textbf{Nouns}: We use the Stanford POS tagger to tag all sentences in the dataset, and find the top 100 nouns (\texttt{NN} tag) in the entire dataset. All words among the top 100 nouns are then masked. This affects $\approx 17.5\%$ of all words in the corpus.
    \item \textbf{Places}: We use the categories from the Places365 dataset, which describe 365 scene categories. All words in our dataset which correspond to the Places365 categories are subsequently masked. This affects $\approx 5.4\%$ of all words in the corpus.
\end{itemize}

\subsubsection{Visual Features}

We preprocess images as in~\cite{Caglayan2018multimodal}. We experiment with two visual feature types to capture different semantic information from the images:
\begin{itemize}[leftmargin=*]
    \item \textbf{Object Features}: We use a ResNet-50 CNN \cite{DBLP:conf/cvpr/HeZRS16} trained on ImageNet for recognizing 1000 object categories. We extract 2048-dim average pooled features from the penultimate CNN layer.
    \item \textbf{Place Features}: A ResNet-50 trained on Places365 \cite{zhou2017places} for scene recognition with 365 categories. We extract the posterior class probabilities, which gives us 365-dimensional visual features.
\end{itemize}

\subsubsection{Acoustic Features}

We extract 40-dimensional filter bank features from 16kHz raw speech signals using a time window of 25ms and an overlap of 10ms, using Kaldi~\cite{Povey_ASRU2011}, and 3-dimensional pitch features. To corrupt the original audio signals, we first extract word-audio alignments from a Kaldi GMM-HMM model trained on the Wall-Street Journal Corpus\footnote{By manually inspecting the alignments, we concluded that their quality was good enough for the purpose of this paper.}, and then we increase the beginning and end timing marks by 25\% of the segment duration to account for possible misalignments. Finally, we use SoX\footnote{\url{http://sox.sourceforge.net/}} and the audio-word alignments to mask the words with white noise or silence. To remove a possible bias induced by the duration of the masked audio segment, the silence/white noise masks are of fixed duration of 0.5 seconds.

\subsubsection{Model Implementation}

The model hyper parameters are the same ones as in~\cite{srinivasan}. All models are implemented using the nmtpytorch framework~\cite{DBLP:journals/pbml/CaglayanGBABB17}. For each experiment, we train three models with different random seeds and report the average results.

\subsection{Evaluation Metrics}





For consistency with previous ASR work, we report Word Error Rate (WER). Because WER is sensitive to variations in unmasked words, it is not a direct indicator of the ability to recover masked words. We propose a second metric, called \textbf{Recovery Rate} (RR), which evaluates the models' ability to accurately recover masked-words.
\begin{align*}
    \text{Recovery Rate} = \frac{\# \text{correctly transcribed masked words}}{\# \text{masked words in test set}} \times 100\%
\end{align*}

To compute our metrics, we use sclite\footnote{\url{http://www1.icsi.berkeley.edu/Speech/docs/sctk-1.2/sclite.htm}} which provides WER and hypothesis-reference word alignments for RR calculation.

\section{Results \& Analysis}
\label{sec:analysis}
\subsection{Results}
\label{subsec:results}

In Table 1, we summarize the results of our masking experiments. We see that multimodal ASR models consistently improve over unimodal ones on WER, under all masking conditions. 
We also note that Object Features consistently outperform the Place Features, even when place words are masked. We hypothesize that this might be due to a domain mismatch between Places365 and Flickr8k datasets. It is important to note that due to an intrinsic unbalance between nouns and places in Flickr8k, the results on place and noun masking are not comparable. We leave a more accurate comparison for future work.


\begin{table}[b]
\centering
\begin{tabular}{|c|c|c|c|}
\hline
Masked Words            & Model           & WER & RR \\ \hline \hline
\multirow{3}{*}{Nouns}  & Unimodal        &  32.9\%  & 46.0\%   \\ \cline{2-4} 
                        & Multimodal (Object)        &  29.9\%  & 57.8\%   \\ \cline{2-4}
                        & Multimodal (GT) & 25.3\%     & 92.63\%   \\ \hline \hline
\multirow{3}{*}{Places} & Unimodal        & 22.6\%    & 42.7\%   \\ \cline{2-4} 
                        & Multimodal (Object)        &  19.3\%   & 57.3\% \\ \cline{2-4}
                        & Multimodal (GT) &  20.5\%   & 92.66\%   \\ \hline
\end{tabular}
\caption{Comparison of multimodal improvements when we provide ground-truth (GT) information about what words have been masked}
\label{tab:upper-bound}
\end{table}

However, previous work has suggested that visual modality is only being used as a regularization signal~\cite{Caglayan2018multimodal}. To investigate this behavior, we also measure the models' ability to recover masked words - ensuring that the observed improvements come from the semantics of the visual context. We observe that multimodal models have a significant improvement in noun and place RR, with both Object and Place features. Once again, Object features seem to be more useful for recovery than Place features, even when place words are masked. However, we also note that Place features are better at recovering place words than nouns. These improvements in RR demonstrate that visual context is highly useful when it comes to recovering information which is missing in the primary modality.




\begin{table*}[t]
\label{tab:examples}
\centering
\begin{tabular}{p{3cm} l l}

     \midrule
     \multirow{5}{*}{\includegraphics[width=3cm]{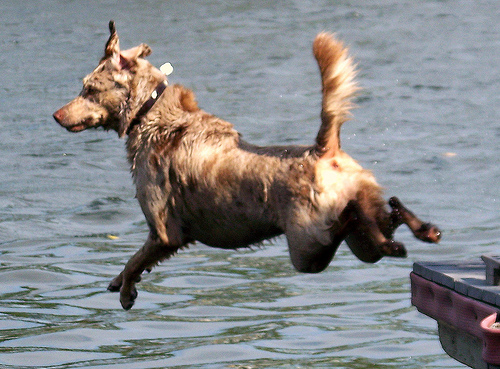}} & \vspace{-2mm} \\
     & Reference & A \textcolor{blue}{dog} wearing a collar jumping from a platform . \\
     & LM & A \textcolor{red}{man} is a red is over a blue \\
     & Unimodal ASR & A \textcolor{red}{woman} wearing a collar jumping into a hurdle . \\
     & Multimodal (object feats) & A \textcolor{ForestGreen}{dog} wearing a collar jumping through a puddle . \\
     & Multimodal (place feats) & A \textcolor{ForestGreen}{dog} wearing a collar jumping over a hurdle.
     \vspace{2.5mm} \\
     \midrule
     \multirow{5}{*}{\includegraphics[width=3cm]{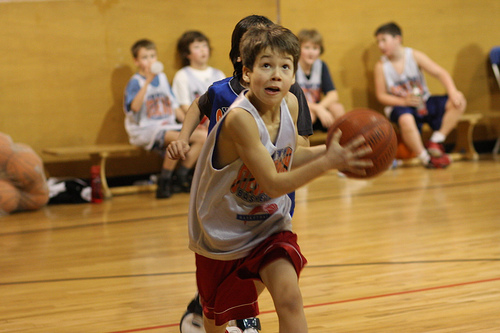}} & \vspace{-3mm} \\
     & Reference & A \textcolor{blue}{boy} with red shorts is holding a \textcolor{blue}{basketball} in a \textcolor{blue}{basketball} court . \\ 
     & LM & Man \textcolor{red}{in} a hair is jumping a \textcolor{red}{stick} . a field \textcolor{red}{ball} \\
     & Unimodal ASR & A \textcolor{ForestGreen}{boy} with red shorts is holding a \textcolor{red}{soccer} ball . \\ 
     & Multimodal (object feats) & A \textcolor{red}{man} in red shorts is holding a \textcolor{ForestGreen}{basketball} in a \textcolor{ForestGreen}{basketball} court . \\
     & Multimodal (place feats) & A \textcolor{red}{man} with red shorts is holding a \textcolor{ForestGreen}{basketball} in a \textcolor{ForestGreen}{basketball} court .
     \vspace{2mm}
     \\
     \hline
     
\end{tabular}
\caption{Examples of noun recovery from the different models. \textcolor{blue}{Blue} words indicate reference words (nouns or places) which are masked with silence in the input speech signal, \textcolor{red}{red} and \textcolor{ForestGreen}{green} words indicate incorrectly and correctly substituted words in the hypotheses transcriptions. The language model (LM) outputs are generated by feeding in the ground-truth reference context at each timestep (not the generated context). }
\end{table*}


An interesting observation is that even with no visual information, a unimodal ASR model is able to successfully recover $\approx 45\%$ of the masked words. We hypothesize that this is due to the predictable nature of the captions in Flickr 8k (similar to observations in Multi30k \cite{caglayan2019probing}). Due to this structured nature of the text, the decoder's implicit language model is able to predict the correct masked word using the decoded text context. To test this hypothesis, we train a GRU language model (LM) and calculate the masked word prediction accuracy when it is fed the ground-truth context. The LM correctly predicts the masked word 34\% of the time. Because the decoder learns to sample from the set of masked words when it encounters silence, we only consider the instances where the LM makes a prediction from the set of masked words. We observe that the LM masked word accuracy jumps to 45\%, matching the unimodal ASR RR. We conclude that the high RR for the unimodal system is a consequence of the structured captions in Flickr 8k, and the subsequent strong LM.

While our multimodal models show significant improvements in WER and RR, we believe that there is a lot of room for improvement. To set an upper bound, we provide a binary vector that encodes which words have been masked. In Table~\ref{tab:upper-bound}, we summarize the improvements in WER and RR obtained by feeding ground-truth masking information when nouns and places are masked with silence. These results encourage further research on more semantically meaningful features for this task. We leave this investigation for future work.


\subsection{Congruency Analysis}
\label{subsec:cong}
In Table \ref{tab:incongruency}, we summarize the results of our congruency experiments (Section \ref{subsec:congruency-exps}).
When incorrect visual features are provided during inference time (\textit{Incongruent Decoding}), the models' performance drops significantly on both the WER and RR metrics. 
These results show that incorrect visual context is actively harming the model. We can conclude that the model learns some useful information during training which is not present during test time.

\begin{table}[h!]
\begin{tabular}{|c|c|c|c|}
\hline
Masked Words & Visual Features & Rel RR $\triangle$ & Rel WER $\triangle$ \\ \hline \hline
\multicolumn{4}{|c|}{Incongruent Decoding} \\    \hline 
\multirow{2}{*}{Nouns}        & Congruent       & 25.5\%                           & 9.1\%                             \\ \cline{2-4}
        & Incongruent     & -43.5\%                           & -28.3\%                            \\ \hline 
\multirow{2}{*}{Places}       & Congruent       & 33.5\%                           & 14.6\%                              \\ \cline{2-4}
      & Incongruent     & -52.7\%                          & -17.4\%                            \\ \hline \hline
\multicolumn{4}{|c|}{Incongruent Training} \\ \hline
\multirow{2}{*}{Nouns}        & Congruent       & 25.5\%                           & 9.1\%                             \\ \cline{2-4}
        & Incongruent     & 2.8\%                           & 6.2\%                            \\ \hline
\multirow{2}{*}{Places}       & Congruent       & 33.5\%                           & 14.6\%                              \\ \cline{2-4}
        & Incongruent     & 3.1\%                          & 10.9\%                            \\ \hline 

\end{tabular}
\caption{Comparison of multimodal models' performance under the incongruent decoding and training setups. Object features and silence masking were used for all experiments.}
\label{tab:incongruency}
\end{table}



When we train the models using misaligned images (\textit{Incongruent Training}), we don't observe a considerable improvement over unimodal ASR on the masked RR task. We believe that the slight RR gains are due to semantic overlap between images. We also obtain larger improvements in WER (not as much as congruent training), consistent with the regularization effect previously observed in \cite{Caglayan2018multimodal}.




\subsection{Qualitative Analysis}
\label{subsec:examples}

In Table 3, we consider several examples in the test set, and compare how unimodal and multimodal models transcribe. In the first example, the LM and unimodal ASR substitute words that occur frequently in the dataset (\textit{i.e.} man and woman, repectively). However, the multimodal models correctly recover the word ``dog''.


In the second example, unlike the multimodal systems, the unimodal model recovers the word ``boy'' correctly. However, the multimodal models, relying on the image, identifies the object (\textit{i.e.} basketball) and scene (\textit{i.e.} basketball court) and correctly recover the masked words. They are able to reason that the masked word is something the boy is holding, and localize the basketball correctly.

\section{Conclusions and Future Work}
\label{sec:conc}
We show that when we deterministically mask out words in the input speech signal using silence/white noise, having auxiliary context from images helps us improve recovery of masked words, demonstrating that the model can avail semantic information from the image.

For future work, we plan to extend the structured experiments (\textit{i.e.} masking concrete words) performed in this paper into an unstructured regime (\textit{i.e.} speech corrupted by overlapping noise), to see if multimodality can help with speech recognition under noisy conditions. If the results are successful, this work could potentially be tested in real-world multimodal environments such as~\cite{sanabria18how2}.


\section{Acknowledgments}
\label{sec:ack}
This research was supported in part by DARPA grant FA8750-18-2-0018 funded under the AIDA program, along with faculty research grants from Amazon, AWS and Facebook. 
This work used the computational resources of the PSC Bridges cluster at Extreme Science and Engineering Discovery Environment (XSEDE)~\cite{6866038}. 
We would also like to thank the anonymous reviewers, along with Shikib Mehri for their valuable inputs.


\bibliographystyle{IEEEbib}
\bibliography{strings,refs}

\end{document}